\documentclass[a4paper,conference]{IEEEtran}
\IEEEoverridecommandlockouts

\usepackage[left=1.57cm,right=1.57cm,top=0.95cm,bottom=2.54cm]{geometry}

\usepackage[utf8]{inputenc} 
\usepackage[T1]{fontenc}

\usepackage{cite}

\ifCLASSINFOpdf
  \usepackage[pdftex]{graphicx}

\else

\fi

\usepackage{epstopdf}
\usepackage{epsfig}

\usepackage[cmex10]{amsmath}

\usepackage{multirow}
\usepackage{array}
\usepackage[lofdepth,lotdepth]{subfig}

\graphicspath{ {./images/} }

\begin{document}


%
\title{Kullanıcı Profili ve İşbirlikçi Filtreleme Tabanlı\\Otel Öneri Sistemi\\ {\fontsize{16}{20}\selectfont Hotel Recommendation System Based on User Profiles \\ and Collaborative Filtering}}

\author{\IEEEauthorblockN{Bekir Berker Türker}
\IEEEauthorblockA{\textit{Veri Bilimi ve Analitiği Bölümü } \\
\textit{Etstur}\\
İstanbul, Türkiye \\
berker.turker@etstur.com}
\and

\IEEEauthorblockN{Resul Tugay}
\IEEEauthorblockA{\textit{Bilgisayar ve Bilişim Fakültesi} \\
\textit{İstanbul Teknik Üniversitesi}\\
İstanbul, Türkiye \\
tugayr@itu.edu.tr}

\\
\IEEEauthorblockN{Şule Öğüdücü}
\IEEEauthorblockA{\textit{Bilgisayar ve Bilişim Fakültesi} \\
\textit{İstanbul Teknik Üniversitesi}\\
İstanbul, Türkiye \\
sgunduz@itu.edu.tr}
\and

\IEEEauthorblockN{İpek Kızıl}
\IEEEauthorblockA{\textit{Bilgisayar Mühendisliği Bölümü} \\
\textit{Koç Üniversitesi}\\
İstanbul, Türkiye \\
ikizil15@ku.edu.tr}

}



\maketitle

\begin{ozet}
Günümüzde, çevrimiçi rezervasyon sistemleri sundukları sayısız imkanlarla acentaların yerini almaktadır ve birçok insan gezi planlarını çevrimiçi rezervasyon sistemleri üzerinden yapmaktadır. Ancak insanlar ne zaman, nereyi ziyaret edeceğine karar vermek için çok fazla zaman harcamak zorunda kalmaktadır. Ayrıca, çevrimiçi sistemlerdeki bu sayısız olasılıklar, kullanıcıların diğerlerinden daha ilginç teklifleri ayırt etmelerini zorlaştırmakta ve daha cazip teklifler farkedilmeden gidebilmektedir. Bu bağlamda öneri sistemleri, insanların karar verme sürecini kolaylaştırabilecek en iyi sistemlerdir. Geleneksel olarak öneri sistemleri içerik tabanlı ve işbirlikçi filtreleme yaklaşımları kullanarak kullanıcıların en uygun ürün ve hizmetleri bulmalarına yardımcı olur. Bu her iki yöntemin dezavantajları olsa da, her ikisinin birleştirilmesiyle oluşturulacak yeni bir melez sistemle öneri sisteminin kalitesi artırılabilir. 
Bu çalışmada, müşterilerin gereksinimi olan oteli en uygun şekilde öneren ve onları zaman kaybından kurtaran içerik tabanlı ve işbirlikçi filtreleme yaklaşımlarının birleştirilmesiyle yeni bir melez otel öneri sistemi geliştirilmiştir. 

\end{ozet}

\begin{IEEEanahtar}
öneri sistemleri, otel önerileri, melez model, işbirlikçi filtreleme, içerik filtreleme.
\end{IEEEanahtar}

\begin{abstract}
Nowadays, people start to use online reservation systems to plan their vacations since they have vast amount of choices available. Selecting when and where to go from this large-scale options is getting harder. In addition, sometimes consumers can miss the better options  due to the wealth of information to be found on the online reservation systems.  In this sense, personalized services such as recommender systems play a crucial role in decision making. Two traditional recommendation techniques are content-based and collaborative filtering. While both methods have their advantages, they also have certain disadvantages, some of which can be solved by combining both techniques to improve the quality of the recommendation. The resulting system is known as a hybrid recommender system. This paper presents a new hybrid hotel recommendation system that has been developed by combining content-based and collaborative filtering approaches that recommends customer the hotel they need and save them from time loss.

\end{abstract}


\section{G{\footnotesize İ}r{\footnotesize İ}ş ve Öncek{\footnotesize İ} Çalışmalar}

Teknolojideki gelişmelerle birlikte dünya çapındaki sayısız ürün, restoran, haber, kitap, film, müzik ve oteller artık çevrimiçi sistemler aracılığıyla ulaşılabilir hale gelmiştir. Ancak tüm bu çeşitlilik beraberinde bu seçenekleri gözden geçirerek aradan en uygun olanı seçme zorluğunu da kendisiyle getirmiş ve bu durum sonucunda kişiselleştirilmiş sistemlere diğer bir adıyla öneri sistemlerine ihtiyaç artmıştır. Öneri sistemlerinde amaç kullanıcıların tercihlerini öğrenmek ve onlara ihtiyaçları doğrultusunda önerilerde bulunmaktır. Genel olarak bahsetmek gerekirse, bir öneri sistemi, belirli bir alandaki öğeler (ürünler veya eylemler) hakkında belirli bir aktif kullanıcının ilgisini çekebileceği düşünülen özel öneriler sunmaktadır.
Öneri sistemleri, makine öğrenmesi, yaklaşım teorisi ve çeşitli sezgisel yöntemler gibi birçok farklı yaklaşım kullanılarak geliştirilebilir. Kullanılan teknikten bağımsız ve önerilerin nasıl yapıldığına bağlı olarak, öneri sistemleri genellikle \cite{Adomavicius} aşağıdaki kategorilere ayrılmaktadır: (1) işbirlikçi filtrelemeye dayalı sistemler; kullanıcının zevkine benzer zevkleri olan insan gruplarını belirlemeye çalışan ve ortak beğenilen öğeleri öneren, (2) içerik tabanlı öneri sistemleri; daha önce kullanıcı tarafından tercih edilenlere benzer öğeler önermek için içerik bilgisini kullanan sistemler olmak üzere iki şekilde sınıflandırılabilir.

\begin{figure*}[t]
	\centering
	\includegraphics[width=0.90\textwidth]{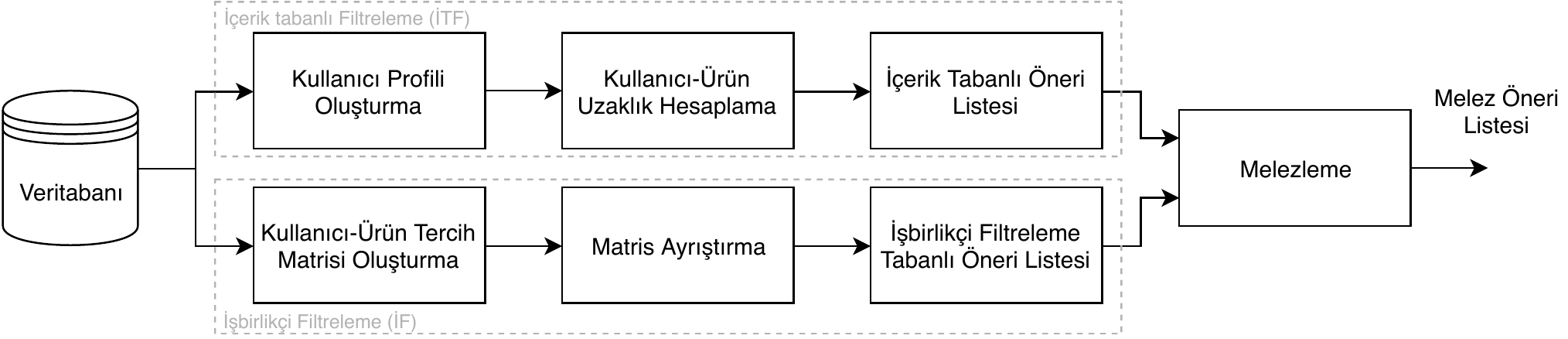}
	\caption{Genel sistem mimarisi}
	\label{fig:sysdiag}
\end{figure*}

Genel olarak, işbirlikçi filtrelemeye dayalı öneri sistemleri içerik tabanlı sistemlerden daha iyi performans göstermektedir, ancak başarıları yeterli sayıda ürün değerlendirmesi varlığına dayanmaktadır \cite{Chen}. Bu tür sistemlerde soğuk başlatma olarak adlandırılan yani yeni bir ürün geldiğinde bu ürün için herhangi bir kayıt bulunmadığından bu ürünü önerememe problemi oluşmaktadır. Bu gibi sorunların çıkma sebebi benzerlik analizinin doğru bir şekilde yapılamamasıdır \cite{CFProblems}. Bu durumlarda içeriğe dayalı bir yaklaşımın kullanılması alternatif olarak ortaya çıkmaktadır \cite{contentbased}. Ancak bu sistemlerin de kendi içerisinde sınırlamaları vardır. Örneğin, öğelerin içeriğini temsil etmek için kullanılan öznitelikler ürünü temsil etmek için yeterli olmayabilir. Ayrıca, içerik tabanlı yaklaşımlarda çok az değerlendirmesi bulunan kullanıcılara doğru ve çeşitli önerilerde bulunmak zorlaşmaktadır. 

Yukarıda saydığımız problemleri çözecek ortak bir yaklaşım işbirlikçi filtreleme (İF) ve içerik tabanlı filtreleme (İTF) sistemleri bir araya getirilerek yeni bir melez sistem oluşturulmasıdır \cite{BurkeHybrid}. İki farklı öneri sistemi birleştirilirken ağırlıklı kriter kullanımı (farklı öneri sistemlerinden çıkan skorlar sayısal olarak birleştirilir), anahtarlama mekanizmasının kullanımı (sistem herhangi bir öneri sisteminden önerilecek ürünü/hizmeti seçer ve önerir) şeklinde farklı melezleme teknikleri kullanılarak yapılabilmektedir. Hatta iki öneri sistemi aynı anda kullanılabilmektedir, bunun nasıl olacağının seçimi tamamen öneri sistemini geliştiren kişiye bağlıdır. Melez bir sistemin nasıl oluşturulacağı ve bununla ilgili kriterlerin belirlenmesi melezleme sırasında çözülmesi gereken yaygın sorunlardan biridir.

Öneri sistemleri üzerine literatürde bolca çalışma bulmak mümkündür. Bu çalışmaların nispeten büyük çoğunluğu da müzik \cite{su2010music,yoshii2008music}, film \cite{diao2014movie,said2010movie,lekakos2008movie} ve e-ticaret ürünlerinin \cite{schafer2001commerce,wang2013commerce,Amazon} kullanıcı etkileşimleri ile öneri oluşturma üzerinedir. Bu sistemleri Spotify, Netflix, Amazon gibi firmalar aktif olarak kullanmakta ve kullanıcıları için önemli ölçüde faydalar sağlamaktadırlar.

Ayrıca, öneri sistemleri için literatür taraması sunan çalışmalar bulunmaktadır. Ricci ve diğerleri, çalışmalarında \cite{ricci2015recommender} öneri sistemlerini genel olarak ele almış ve bu alandaki gerekli tanımlama, sınıflandırma, kullanım alanları gibi bilgileri sunmuşlardır. Benzer şekilde ve nispeten yakın zamanda Mu ve diğerleri, derin öğrenme tabanlı öneri sistemlerinin literatür taramasını sunan bir çalışma \cite{mu2018deep} gerçekleştirmişlerdir. Bu çalışmada, otomatik kodlayıcı, kısıtlı Boltzmann makineleri, özyinelemeli sinirsel ağlar, çekişmeli üretici ağlar gibi sinir ağları yapıları ile mümkün kılınan tekil ve melez öneri sistemleri ele alınmıştır. Bir başka çalışmada \cite{chen2018cf} Chen ve diğerleri, işbirlikçi filtreleme yöntemi özelinde öneri sistemlerinin literatür taramasını sunmuştur. Bu konuda kullanılan geleneksel yöntemler kısıtları ile birlikte verilmiş ve güncel çözüm yöntemlerinden bahsedilmiştir.

Turizm alanında ise rota, etkinlik, otel gibi önerilerin yapıldığı bir takım çalışmalar mevcuttur. Örneğin, Fang-Ming ve diğerleri çalışmalarında \cite{fangming2012attractions}, Bayes ağları kullanarak ilgili lokasyonlarda kişiye özel turistik etkinlikler önerileri oluşturmuşlar ve Google haritalar ile entegre etmişlerdir. Benzer şekilde, kullanıcının kısıtlarını ve isteklerini de göz önünde bulundurarak zaman ve masraf açısından en uygun rotayı öneren çalışmalar da mevcuttur \cite{liu2015route}. 

Otel öneri sistemlerinde genel olarak değinilen bir problem de etkileşim verisinin azlığı ve dolayısıyla etkileşim matrisinin oldukça seyrek bir yapıda olmasıdır. Bu durum literatürdeki çoğu çalışmada adreslenmiş ve çözüme yönelik yöntemler önerilmiştir. Saga ve diğerleri, otel önerisi problemini kullanıcıların yalnızca otel tercihleri verisine dayandırarak geçiş ağları ile modellemiştir \cite{saga2008hotel}. Otel tercihlerinin yanı sıra veri temsilini zenginleştirmek adına ve seyreklik problemine çözüm olarak misafir görüşlerini de ele alan çalışmalar mevcuttur

\cite{takuma2016review,jalan2017context}. Benzer bir amaç ve soğuk başlatma problemine yönelik olarak Qi ve diğerleri, uçuş tercihleri ile otel tercihleri arasında dönüşüm matrisi kullanarak kullanıcı karakterini zenginleştirdiklerini ifade etmişlerdir \cite{qi2018cross}. Zhang ve diğerleri ise, çalışmalarında \cite{zhang2015hotel} kullanıcının önceki tercihleri ve geçerli zamandaki niyetlerini kullanarak melez bir sistem oluşturmuşlardır.

Bu çalışmada otel öneri sisteminin gerçek hayattaki performansını arttırmak için içerik tabanlı ve işbirlikçi filtrelemeye dayalı iki öneri sistemi bir araya getirilerek melez bir öneri sistemi geliştirilmiştir. 
Literatürdeki benzer çalışmalardan farklı olarak bu çalışma, içerik tabanlı yöntemde zengin bir otel öznitelik listesinden faydalanmakta ve işbirlikçi filtreleme kısmında etkileşim matrisini misafirlerin tercih miktarları ile oluşturarak kullanıcı-ürün skor matrisine benzetmektedir. Son olarak anahtarlama yöntemi ile iki ayrı sistemi melezleyerek çıktı oluşturmaktadır.

Yapılan çalışmanın detaylarını sunmak için bu makale aşağıdaki şekilde yapılandırılmıştır. Bir sonraki bölüm, geliştirilen otel öneri sistemi ve bu sistem kurulurken kullanılan yaklaşımları detaylı bir şekilde tanıtmaktadır. Önerilen modelin performansı Bölüm \ref{sec:deneyler}'te değerlendirilmiştir. Son olarak, Bölüm \ref{sec:vargilar} sonuçlarımızı genel hatlarıyla özetlemektedir.

\IEEEpubidadjcol

\section{Otel Öner{\footnotesize İ} S{\footnotesize İ}stem{\footnotesize İ}}
\label{sec:otelOneriSis}

Öneri sistemleri oluşturmak için bir çok yöntem bulunmaktadır. Fakat hepsinin esas amacı her kullanıcı için ürün kümesindeki elemanlara karşılık gelen skor/değerlendirme değerlerini üretmektir. Bu çıktıları üretirken kullanıcıların daha önceden belli ürünler için sağlamış olduğu skor/değerlendirme değerleri kullanılmaktadır. Kullanıcıya özel öneri listesi oluşturulurken ise bu ürünler liste dışında tutulmaktadır.
Buradaki amaç müşterinin daha önce etkileşimde bulunmadığı ürünlerden birini önerebilmektir. Sonuç olarak, hesaplanan skor/değerlendirme sonuçları sıralanarak liste başından belli sayıda ürün seçilmekte ve müşteriye sunulmaktadır. 

Otel öneri sistemimizde benzer yöntemler takip edilerek iki ayrı koldan iki farklı öneri listesi oluşturularak melez bir öneri listesi elde edilmektedir. Şekil \ref{fig:sysdiag}'de sistemin genel mimarisi verilmiştir. Burada, üst kolda İTF tabanlı yöntem akışı bulunurken alt kolda ise İF tabanlı yöntem akışı bulunmaktadır ve son olarak tekil çıktılar birleştirilerek melez çıktı elde edilmektedir. Takip eden başlıklarda ilgili alt sistem ve basamakların detayları verilmiştir.


\subsection{İçerik Tabanlı Filtreleme Yöntemi}
\begin{figure}[h]
	\centering
	\includegraphics[width=0.30\textwidth]{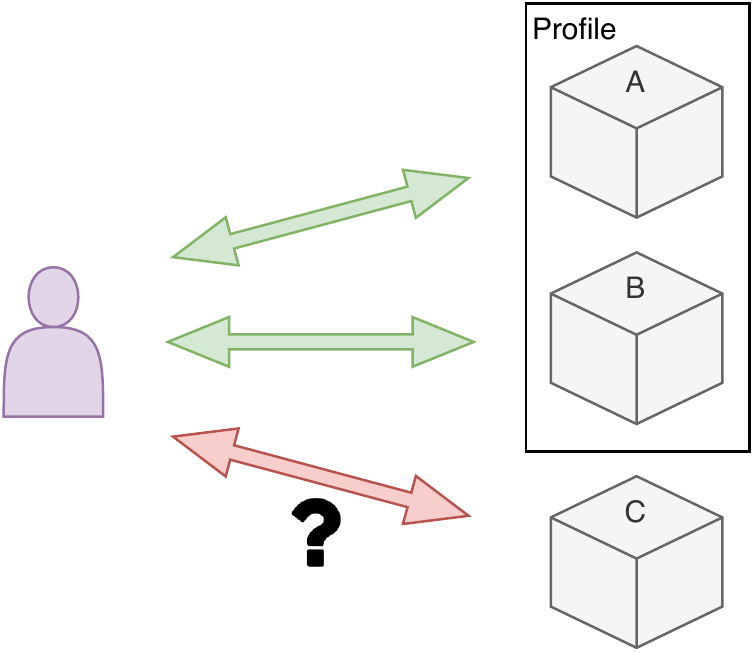}
	\caption{İçerik tabanlı filtreleme yöntemi}
	\label{fig:itf}
\end{figure}

İTF yönteminde kullanıcılar temel olarak sahip oldukları veya etkileşime geçtikleri ürünlerin özellikleri ile temsil edilmektedirler. 
Bu durum kitap satın alan müşteriler için satın alınan kitabın türü, ücreti, sayfa sayısı gibi özellikler olurken film izleyen kullanıcılar için film ile ilgili özellikler olmaktadır. Kullanıcıları kullandıkları ya da satın aldıkları ürünlerin özellikleri ile temsil etmek aslında onların profillerini çıkartmaktır.
Şekil \ref{fig:itf}'de görülen kullanıcı daha önce A ve B ürünleri ile ilişkilendirilmiş ve bu iki ürün kullanıcının profilinin oluşmasını sağlamıştır. Oluşturulan bu profil ışığında, kullanıcının C ürünü ile etkileşiminin olasılığı sorgulanabilmektedir.
Bu çalışmada da kullanıcıların profillerini çıkartmak için daha önce gittikleri otellerin özellikleri esas alınmıştır.

Örneğin Tablo \ref{content}'de kullanıcı \textit{1}, $\textit{A}$ ve $\textit{B}$ otellerine gitmiş olsun. \textit{Kullanıcı 1}'in gittiği otellerin kapasiteleri sırasıyla 500 ve 700 kişilik, denize olan uzaklıkları 150 ve 50 metre ve bu iki otelde de kahvaltı hizmeti olsun. Dolayısı ile \textit{Kullanıcı 1}'in profili hesaplanırken bu içerik bilgilerinden yararlanılır ve Tablo \ref{content}'deki ilgili satırların ortalama değerleri alınır. Bu durumda \textit{Kullanıcı 1}'in gittiği otellerin kapasitesi ortalama olarak 600 kişilik, denize uzaklıkları ise 100 metredir. Ayrıca \textit{Kullanıcı 1} gittiği otellerde kahvaltı hizmeti de aramaktadır. Diğer taraftan \textit{Kullanıcı 2} daha az kapasiteli otelleri ve denize daha uzak otelleri seçerek daha farklı bir profil sergilemiştir. Sonuç olarak bu kullanıcıların profilleri Tablo \ref{content-result}'deki gibi olur.

\begin{table}[h]
\caption{Kullanıcı-Otel Bilgileri Örneği}
\label{content}
\centering
\begin{tabular}{|c|c|c|c|c|}
\hline
\multicolumn{1}{|l|}{\textbf{Kullanıcı}} & \multicolumn{1}{l|}{\textbf{Otel}} & \textbf{Kapasite} & \textbf{DenizeOlanUzaklık} & \textbf{Kahvaltı} \\ \hline
1                                        & A                                  & 500             & 150            & 1             \\ \hline
1                                        & B                                  & 700             & 50             & 1             \\ \hline
2                                        & C                                  & 200             & 500            & 0             \\ \hline
2                                        & D                                  & 300             & 1000           & 0             \\ \hline
2                                        & E                                  & 100             & 1500           & 1             \\ \hline
\end{tabular}
\end{table}

Kullanıcıların profilleri ile sistemde var olan fakat kullanıcının gitmediği oteller arasında benzerlik hesaplanarak kullanıcıların profiline en yakın $N$ tane otel önerilmiştir. Otel listesi veri ön işleme adımlarından geçirilerek temizlenmiştir. Bu ön işleme adımları, eksik verileri düzeltme (Ör: Otel isimleri),  tutarsız verileri kaldırma (Ör: Oda sayısı) gibi işlemlerden oluşmaktadır. Daha sonra otellerin en önemli 11 özelliği Temel Bileşenler Analizi (TBA) yöntemi kullanılarak 220 tane özellik arasından seçilmiştir. Bu özelliklerin hepsi burada listelenemeyeceği için birkaç tanesinden bahsedilmiştir. Bunlar: otelin fiyatı, otelin şehir merkezine olan uzaklığı, otelde otopark olup olmaması vb. TBA yöntemi kullanılarak veri boyutunun azaltılması  
sayesinde profiller ile oteller arasında benzerlik hesaplaması yapılırken çalışma süresinin kısalması sağlanmıştır.

\begin{table}[h]

\caption{İTF ile kullanıcıların temsil edilmesi}
\label{content-result}
\centering
\begin{tabular}{|c|c|c|c|}
\hline
\textbf{Kullanıcı} & \textbf{Kapasite} & \textbf{DenizeOlanUzaklık} & \textbf{Kahvaltı} \\ \hline
1                  & 600            & 100                        & 1                 \\ \hline
2                  & 200            & 1000                       & 0,33              \\ \hline
\end{tabular}
\end{table}

\subsection{İşbirlikçi Filtreleme Yöntemi}
\label{sec:CF}
\begin{figure}[h]
	\centering
	\includegraphics[width=0.40\textwidth]{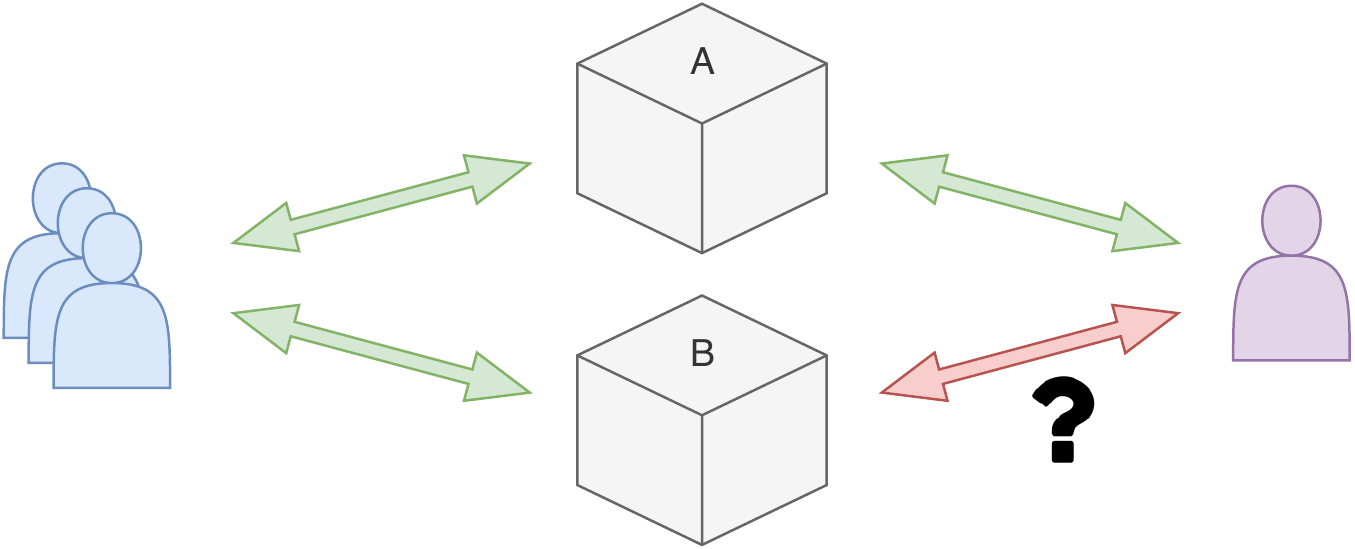}
	\caption{İşbirlikçi filtreleme yöntemi}
	\label{fig:if}
\end{figure}

İşbirlikçi filtreleme kişiselleştirilmiş öneri sistemlerinde kullanılan en önemli yaklaşımlardan birisidir. Kullanıcılar için izole profiller oluşturmak ve bunları ürünler ile ilişkilendirmenin ötesinde kullanıcılar arasında etkileşim de sağlanmaktadır. Bu etkileşim, kullanıcıların seçimlerinde veya oylamalarında birbirine benzerlikleri olarak açıklanabilir. Bir başka deyişle, kullanıcıların tercih benzerlikleri öneri oluşturmak için kullanılmaktadır. 
Örneğin, Şekil \ref{fig:if}'te görüldüğü üzere bir grup kullanıcı (solda) daha önce ürün A ve B ile ilişkilendirilmiş ve sorgu altındaki kullanıcı (sağda) daha önce sadece A ürünü ile ilişkilendirilmiştir. Bu durumda sorgu altındaki kullanıcı için B ürünü ile ilişki ihtimali sorgulanabilir ve yüksek olması beklenmektedir.

İF için birçok teknik bulunmaktadır \cite{CF}. Bunlardan birisi de Netflix Price \cite{Netflix} ile öneri sistemleri üzerinde büyük etki yaratmış Matris Ayrıştırma yöntemidir. Bu yöntem, bir doğrusal cebir konusu olarak herhangi bir matrisin iki dikdörtgen matrise ayrıştırılması esasına dayanır ve teknik detayları 
bir sonraki bölümde verilmiştir.

\subsubsection{Matris Ayrıştırma}
\label{sec:MF}
 Öneri sistemlerinde eğitim için toplanılan kullanıcı değerlendirme skorları ile bir matris oluşturulur ve bu matris problemin doğası gereği oldukça seyrektir. Matris Ayrıştırma tekniği Tekli Değer Ayrışımı (TDA) tekniğine oldukça benzemekle beraber, TDA'nın sahip olmadığı eksik matris girdisi toleransına sahip bir alternatiftir. Denklem~\ref{eq:MF}'de görüldüğü üzere oluşturulan büyük ve seyrek matris ($r_{mu}$) daha küçük ve yoğun iki matrisin ($p_m, q_u$) çarpımı şeklinde ifade edilmektedir \cite{YehudaMF}. 
 
\begin{equation} 
\begin{split}
r_{mu} = p_m * q_u , \\
\hat r_{mu} = \hat p_m * \hat q_u ,
	\label{eq:MF}
\end{split}
\end{equation}

Burada $r_{mu}$ müşteri ve ürün ikililerinin değerlendirme skorlarını barındıran matrisi, $p_m$ müşterilerin gizli uzay (latent space) temsil matrisini, $q_u$ ise ürünlerin gizli uzay temsil matrisini ifade etmektedir. Ayrıştırılmış matrisler ($p_m$, $q_u$) bir sistem parametresi olan gizli uzay boyutunu ($k$) paylaşmaktadırlar. Matris boyutları, $r, p, q$ için sırasıyla $m*u, m*k, k*u$ şeklinde tanımlanmaktadır.

Denklem~\ref{eq:MF}'de ifade edilen $p, q$ matrisleri Gradyan İnişi, Değişimli En Küçük Kareler (DEKK) yöntemi gibi teknikler ile çözülebilmektedir. Çoğu öneri sistemi problemlerinde $m*u$ sonucunun göreceli olarak büyük çıkmasından dolayı En Küçük Kareler yöntemi tercih edilmektedir \cite{YehudaMF,Pilaszy2010ALS}. İki matrisin kestirimi dışbükey olmayan bir problem olduğu için DEKK yöntemi kullanılmaktadır. Değişimli kelimesi, $p$ ve $q$ matrislerinin değişmeli olarak birinin sabit tutulmasını ve diğerine en küçük kareler yönteminin uygulanmasını ifade etmektedir.

DEKK yönteminin bilinen maliyet fonksiyonuna ek olarak düzenlileştirme parametresi eklenmektedir. Buradaki amaç aşırı uyumlama probleminden kaçınarak daha genel bir uyumlama ile $ \hat r_{mu}$ matrisini oluşturmaktır. Denklem~\ref{eq:MF}'de görüldüğü üzere $ \hat r_{mu}$ matrisi, kestirilen  $ \hat p_{m}$ ve $ \hat q_{u}$ matrislerinin çarpımı ile elde edilmektedir. Böylelikle $r_{mu}$ matrisinde bulunan eksik girdiler doldurularak yoğun $ \hat r_{mu}$ matrisi elde edilmektedir. Doldurulan girdiler müşteriler için henüz değerlendirilmemiş/kullanılmamış ürünlerin önerisini oluşturmaktadır.

\subsection{Melez Model Oluşturma Tekniği}
\label{sec:hybrid_technique}

Tekil öneri sistemlerinin çıktıları melez bir öneri sistemi oluşturmak için kullanılmaktadır. Çıktıların birleştirilmesi için skor ile sıralanmış listeler alınmakta ve aynı sıra indisli ürünler arka arkaya gelecek şekilde yeni bir melez liste oluşturulmaktadır. 

\begin{equation} 
\begin{split}
M^N = A^{N/2} \cup B^{N/2} \\
A^{N/2} \cup B^{N/2}= [ A_1, B_1, A_2, B_2, ... A_{N/2}, B_{N/2} ] ,
	\label{eq:hybrid}
\end{split}
\end{equation}

Denklem~\ref{eq:hybrid}'de melez listenin oluşturulma biçimi gösterilmiştir. Burada, tekil sistemler $A$ ve $B$ olarak tanımlanmış, melez sistem ise $M$ olarak tanımlanmıştır. Alt indisler sistemlerin sıralı liste elemanı indislerini belirtirken üst indisler listenin kendisini uzuğunluğuyla beraber ifade etmektedir. Örneğin, $A$ ve $B$ sistem listelerinin ilk 5'er elemanları ($A^5$, $B^5$) alınarak $M^{10}$ listesi (melez sistemin İlk10'u) oluşturulmaktadır.
 
\section{Deneyler}
\label{sec:deneyler}

\subsection{Veri Seti}

Otel öneri sistemi için turizm sektöründen bir şirketin yaklaşık 7 yıl süreli rezervasyon kayıtları kullanılmıştır. Bu kayıtlar anonim olarak oluşturulmuş müşteri kimlik sözcüğü, otel kodu ve tarih olmak üzere 3 kolondan oluşmaktadır. Her bir kayıt bir müşterinin hangi tarihte ve hangi otelde konakladığı bilgisini içermektedir. Kayıt sayısı yaklaşık olarak 4,5 milyondur. 

Müşteri bazında akıllı tekilleştirme uygulanmış \cite{bayrak2018near} veri setindeki eşsiz müşteri sayısı yaklaşık olarak 1,5 milyon, eşsiz otel sayısı ise 10 bin civarındadır. Her müşterinin en az 2 rezervasyonu ve dolayısıyla 2 kaydı bulunmaktadır. Müşteriler tamamen anonimize edilmiş ve müşteriye ait hiç bir ek bilgi kullanılmamıştır. Bunun yanında, otellerin öznitelikleri kullanılmıştır. Basitçe, bu öznitelikler otellerin müşterilerine sundukları imkanları kapsamaktadır. Müşterilerin otelleri seçmesinde rol oynayan bu özniteliklerden bazıları şu şekilde sıralanabilir: açık havuz, spa, evcil hayvan kabul etme.

\subsubsection{Eğitim-Test Senaryoları}

Veri seti, eğitim ve test kümelerine ayırılarak deneyler gerçekleştirilmiştir. Sunulan tekil ve melez sistemlerin farklı eğitim-test kümeleri üzerindeki performanslarını karşılaştırabilmek için 5 senaryoda eğitim-test bölümlendirmesi yapılmıştır.

\begin{table}[h]
\centering
\caption{Senaryoların oluşturulma biçimi}
\label{tab:scenarios}
\begin{tabular}{|c|c|c|}
\hline
\textbf{Senaryo} & \textbf{Rez. Sayısı Aralığı} & \textbf{Test Kümesi}                           \\ \hline
1                & 3-10                                & Her müşterinin son rezervasyonu                \\ \hline
2                & 2-10                                & Her müşterinin son rezervasyonu                \\ \hline
3                & 2-10                                & Senaryo 1'in test kümesi \\ \hline
4                & 3-5                                 & Her müşterinin son rezervasyonu                \\ \hline
5                & 8-10                                & Her müşterinin son rezervasyonu                \\ \hline
\end{tabular}
\end{table}

Tablo~\ref{tab:scenarios}'te senaryoların hangi filtrelerden geçirildiği ve test kümesinin hangi kural ile ayrıldığı verilmiştir. Rezervasyon sayısının farklı olduğu senaryolarda farklı performans ölçümleri alınması beklenmiştir. Test kümesi 3. senaryo hariç benzer şekilde her müşterinin son rezervasyonu seçilerek ayrılmıştır. Örneğin, senaryo 1'de, rezervasyon sayısı 3 ile 10 olan müşterilerin verisi alınmış ve her müşterinin son rezervasyonu test kümesine atılmıştır. Dolayısıyla test kümesindeki kayıt sayısı ile eşsiz müşteri sayısı eşittir. Tablo~\ref{tab:scenario_data_details}'te her senaryo için eğitim ve test kümelerinin sahip olduğu veri boyutları ile ilgili detaylı bilgi verilmiştir. Eşsiz otel sayısı eğitim ve test kümeleri için aynıdır.

Tasarlanan bu senaryolar ile performans karşılaştırmaları yapılarak bir takım vargılara ulaşılması planlanmıştır. Senaryo 1-2 karşılaştırması ile 2 rezervasyonlu müşterilerin etkisi (minimum rezervasyonlu sayıca büyük grup), senaryo 1-3 karşılaştırması ile yalnızca eğitim kümesine katılan 2 rezervasyonlu müşterilerin etkisi incelenmiştir. Senaryo 4-5 karşılaştırması sayesinde de düşük rezervasyonlu grup ile yüksek rezervasyonlu grubun performans farkı incelenmiştir.

\begin{table}[h]
\centering
\caption{Senaryolara ait veri boyutları}
\label{tab:scenario_data_details}
\begin{tabular}{cc|c|c|c|}
\cline{3-5}
                                       &                     & \multicolumn{2}{c|}{\textbf{Eğitim}}            & \textbf{Test}         \\ \hline
\multicolumn{1}{|c|}{\textbf{Senaryo}} & \textbf{Otel Say.} & \textbf{Kayıt Say.} & \textbf{Müşteri Say.} & \textbf{Kayıt Say.} \\ \hline
\multicolumn{1}{|c|}{1}                & 8.785                & 2.326.623               & 693.547                  & 693.547                \\ \hline
\multicolumn{1}{|c|}{2}                & 9.216                & 3.055.646               & 1.422.570                 & 1.422.570               \\ \hline
\multicolumn{1}{|c|}{3}                & 9,927                & 3.784.669               & 1.422.570                 & 693,547                \\ \hline
\multicolumn{1}{|c|}{4}                & 8.066                & 1.440.985               & 552.303                  & 552.303                \\ \hline
\multicolumn{1}{|c|}{5}                & 5.647                & 405.355                & 52.116                   & 52.116                 \\ \hline
\end{tabular}
\end{table}

\subsection{Öneri Sistemlerinin Otel Rezervasyon Verisine Uygulanması}

İTF yönteminde kullanıcı profilleri ile oteller arasında benzerlik hesaplanırken iki farklı yöntem kullanılmıştır. İlk yöntemde profiller ile tüm oteller karşılaştırılarak öneri kümesi oluşturulmuştur. İkinci yöntemde ise daha önce \cite{pakyurek18cluster} çalışmasında da kullanılan kümeleme yöntemi kullanılmıştır. Bu yöntemde, profiller ile önceden k-ortalama algoritması ile kümelenmiş oteller kümesinden en yakın merkezdeki oteller arasında benzerlik hesaplaması yapılmıştır. Öncelikle oteller k-ortalama algoritması ile kümelenmiş ve her küme içinde barındırdığı otellerin özelliklerinin ortalama değerleri ile temsil edilmişlerdir.. Daha sonra kullanıcı profilleri ile bu kümelerin merkezleri karşılaştırılmıştır, son olarak en yakın kümede kullanıcıya en yakın $N$ tane otel önerilmiştir. Benzerlik hesaplaması yapılırken öklid uzaklığı kullanılmıştır.

İF yönteminde müşterilerin otellere gitme sayıları otellere verilen skor olarak kabul edilmiş ve her senaryo için Denklem~\ref{eq:MF}'deki $r_{mu}$ matrisi oluşturulmuştur. Bu matrislere Bölüm~\ref{sec:MF}'de anlatıldığı üzere Matris Ayrıştırma yöntemi uygulanmıştır. Gizli uzay boyutu ($k$) olarak 20, düzenlileştirme parametresi için ise 0.1 seçilmiştir. Elde edilen $\hat r_{um}$ matrisinde müşteri (satır) bazında düzgeleştirme ve sıralama yapılmıştır. Böylelikle her müşteri için önerilen sıralı otel listesi elde edilmiştir.

\subsection{Tekil ve Melez Modellerin Başarımı}

İTF ve İF yöntemleri kullanılarak otel rezervasyon verisi ile yapılan deneyler sonucu tekil sistemlerin ve melez sistemin başarımları sırasıyla Tablo~\ref{tab:content_result},~\ref{tab:cf_result},~\ref{tab:hybrid_result}'de verilmiştir. Başarım kriteri olarak duyarlılık seçilmiş ve tablolarda yüzdelik olarak raporlanmıştır.

\begin{equation} 
\begin{split}
\textit{Duyarlılık} = \frac{\textit{N'lik listedeki ilgili eleman sayısı}}{\textit{toplam ilgili eleman sayısı}}
	\label{eq:recall}
\end{split}
\end{equation}
Duyarlılık, öneri sistemleri için Denklem~\ref{eq:recall}'teki gibi tanımlanmaktadır. Kısaca duyarlılık ölçüsü, test kayıtlarının hangi oranda öneri listeleri tarafından içerdiği şeklinde hesaplanmaktadır. Başarımlar İlk5, İlk10, İlk100 olarak tanımlanmış farklı uzunluktaki öneri listeleri üzerinden elde edilmiştir. Beklenildiği üzere öneri listesi uzunluğu arttıkça duyarlılık oranı da artacak ya da sabit kalacaktır. 

\begin{table}[h]
\centering
\caption{İçerik Tabanlı Filtreleme başarım tablosu}
\label{tab:content_result}
\begin{tabular}{ccccccc}
\multicolumn{7}{l}{}                                                                                                                                                                                                                                                       \\ \hline
\multicolumn{1}{|l|}{}                 & \multicolumn{3}{c|}{\textbf{\begin{tabular}[c]{@{}c@{}}Kümeleme\\    Tabanlı Veri\end{tabular}}}                     & \multicolumn{3}{c|}{\textbf{Tüm Veri}}                                                                          \\ \hline
\multicolumn{1}{|c|}{\textbf{Senaryo}} & \multicolumn{1}{c|}{\textbf{İlk5}} & \multicolumn{1}{c|}{\textbf{İlk10}} & \multicolumn{1}{c|}{\textbf{İlk100}} & \multicolumn{1}{c|}{\textbf{İlk5}} & \multicolumn{1}{c|}{\textbf{İlk10}} & \multicolumn{1}{c|}{\textbf{İlk100}} \\ \hline
\multicolumn{1}{|c|}{\textbf{1}}       & \multicolumn{1}{c|}{8,17}          & \multicolumn{1}{c|}{10,61}          & \multicolumn{1}{c|}{27,45}           & \multicolumn{1}{c|}{9,62}          & \multicolumn{1}{c|}{12,19}          & \multicolumn{1}{c|}{29,39}           \\ \hline
\multicolumn{1}{|c|}{\textbf{2}}       & \multicolumn{1}{c|}{\textbf{11,05}}         & \multicolumn{1}{c|}{\textbf{12,92}}          & \multicolumn{1}{c|}{26,78}           & \multicolumn{1}{c|}{\textbf{14,10}}         & \multicolumn{1}{c|}{\textbf{16,11}}          & \multicolumn{1}{c|}{29,50}           \\ \hline
\multicolumn{1}{|c|}{\textbf{3}}       & \multicolumn{1}{c|}{8,17}          & \multicolumn{1}{c|}{10,61}          & \multicolumn{1}{c|}{27,45}           & \multicolumn{1}{c|}{9,62}          & \multicolumn{1}{c|}{12,19}          & \multicolumn{1}{c|}{29,39}           \\ \hline
\multicolumn{1}{|c|}{\textbf{4}}       & \multicolumn{1}{c|}{8,42}          & \multicolumn{1}{c|}{10,83}          & \multicolumn{1}{c|}{27,18}           & \multicolumn{1}{c|}{9,76}          & \multicolumn{1}{c|}{12,32}          & \multicolumn{1}{c|}{29,26}           \\ \hline
\multicolumn{1}{|c|}{\textbf{5}}       & \multicolumn{1}{c|}{6,83}          & \multicolumn{1}{c|}{9,35}           & \multicolumn{1}{c|}{\textbf{27,67}}           & \multicolumn{1}{c|}{10,89}         & \multicolumn{1}{c|}{13,34}          & \multicolumn{1}{c|}{\textbf{31,25}}           \\ \hline
\end{tabular}
\end{table}

Tablo~\ref{tab:content_result}'te İTF sonuçları kümelenmiş veriler kullanılarak alınan ve tüm veri kullanarak alınan performansları ayrı ayrı göstermektedir. Tüm veri ile elde edilen sonuçlardaki başarım görece yüksek olmasına rağmen işlem yükü olarak oldukça pahalıdır. Diğer yandan, kümeleme kullanımı beklenildiği üzere başarım kaybına sebep olurken işlem yükü tasarrufu sağlamaktadır.

\begin{table}[h]
\centering
\caption{İşbirlikçi Filtreleme başarım tablosu}
\label{tab:cf_result}
\begin{tabular}{|c|c|c|c|}
\hline
\textbf{Senaryo} & \textbf{İlk5} & \textbf{İlk10} & \textbf{İlk100} \\ \hline
\textbf{1}       & 7.42          & 11.22          & 34.10           \\ \hline
\textbf{2}       & \textbf{7.70}          & 11.42          & 34.17           \\ \hline
\textbf{3}       & 7.57          & \textbf{11.50}          & \textbf{34.93}           \\ \hline
\textbf{4}       & 7.60          & 11.42          & 34.38           \\ \hline
\textbf{5}       & 6.21          & 9.63           & 30.79           \\ \hline
\end{tabular}
\end{table}

Tablo~\ref{tab:cf_result}'da İF sisteminin sonuçları verilmiştir. İTF sonuçları ile karşılaştırıldığında İTF sistemin kısa listede (İlk5) daha başarılı olduğu fakat uzun listede (İlk100) İF yönteminin öne geçtiği gözlemlenmiştir. 

Senaryolar karşılaştırıldığında, İTF yönteminin senaryo 2'deki görece yüksek performansı 2 rezervasyonlu grubu başarılı bir şekilde tahmin ettiğini göstermektedir. Bu grubun sadece eğitime katıldığı 3. senaryoda ise İF sisteminde önemli bir artış görülmemiştir. İTF'de eğitim kümesine eklenen veriler herhangi bir etkide bulunmadığı için senaryo 1 ve 3 performansları Tablo~\ref{tab:content_result}'te aynıdır. Öte yandan, senaryo 4 ve 5 karşılaştırıldığında, her iki tekil sistemde de yüksek rezervasyonlu müşteri verisinde (senaryo 5) nispeten daha düşük performans alındığı gözlemlenmiştir.

İki yöntemde de senaryo 5'in  performansının diğer senaryolara göre nispeten düşük olmasının sebebi bu senaryoya ait veri setinin küçük olması olarak gösterilebilir. Kullanıcılara ait rezervasyon miktarı veri zenginliğini arttırmaktadır fakat yüksek rezervasyonlu müşteri sayısı oldukça düşüktür. Veri seti büyüklüğünün önemi senaryo 2'nin göstermiş olduğu performans artışı ile de ortaya çıkmaktadır. Özellikle İTF yönteminde dramatik bir artış gösteren senaryo 2, İF yönteminde aynı etkiyi gösterememiştir. Bunun sebebi olarak da Matris Ayrıştırma yönteminin az ürün/skor değerlendirme sayılı veriler ile nispeten daha kötü çalışması olarak gösterilebilir.

\begin{table}[h]
\centering
\caption{Melez Filtreleme başarım tablosu}
\label{tab:hybrid_result}
\begin{tabular}{ccccccc}
\multicolumn{7}{l}{}                                                                     \\ \hline
\multicolumn{1}{|l|}{}                 & \multicolumn{3}{c|}{\textbf{\begin{tabular}[c]{@{}c@{}}Kümeleme\\    Tabanlı\end{tabular}}}                     & \multicolumn{3}{c|}{\textbf{Tüm Veri}}                                                                          \\ \hline
\multicolumn{1}{|c|}{\textbf{Senaryo}} & \multicolumn{1}{c|}{\textbf{İlk5}} & \multicolumn{1}{c|}{\textbf{İlk10}} & \multicolumn{1}{c|}{\textbf{İlk100}} & \multicolumn{1}{c|}{\textbf{İlk5}} & \multicolumn{1}{c|}{\textbf{İlk10}} & \multicolumn{1}{c|}{\textbf{İlk100}} \\ \hline
\multicolumn{1}{|c|}{\textbf{1}}       & \multicolumn{1}{c|}{10.62}         & \multicolumn{1}{c|}{15.25}          & \multicolumn{1}{c|}{41.81}           & \multicolumn{1}{c|}{12.01}         & \multicolumn{1}{c|}{16.73}          & \multicolumn{1}{c|}{44.09}           \\ \hline
\multicolumn{1}{|c|}{\textbf{2}}       & \multicolumn{1}{c|}{\textbf{14.07}}         & \multicolumn{1}{c|}{\textbf{18.31}}          & \multicolumn{1}{c|}{42.21}           & \multicolumn{1}{c|}{\textbf{17.16}}         & \multicolumn{1}{c|}{\textbf{21.46}}          & \multicolumn{1}{c|}{\textbf{45.72}}           \\ \hline
\multicolumn{1}{|c|}{\textbf{3}}       & \multicolumn{1}{c|}{10.69}         & \multicolumn{1}{c|}{15.42}          & \multicolumn{1}{c|}{\textbf{42.43}}           & \multicolumn{1}{c|}{12.07}         & \multicolumn{1}{c|}{16.92}          & \multicolumn{1}{c|}{44.70}           \\ \hline
\multicolumn{1}{|c|}{\textbf{4}}       & \multicolumn{1}{c|}{11.09}         & \multicolumn{1}{c|}{15.69}          & \multicolumn{1}{c|}{41.88}           & \multicolumn{1}{c|}{12.25}         & \multicolumn{1}{c|}{17.05}          & \multicolumn{1}{c|}{44.20}           \\ \hline
\multicolumn{1}{|c|}{\textbf{5}}       & \multicolumn{1}{c|}{8.43}          & \multicolumn{1}{c|}{12.88}          & \multicolumn{1}{c|}{39.21}           & \multicolumn{1}{c|}{12.77}         & \multicolumn{1}{c|}{17.18}          & \multicolumn{1}{c|}{43.36}           \\ \hline
\end{tabular}
\end{table}

Tablo~\ref{tab:hybrid_result}'de performansları verilen melez öneri listeleri Bölüm~\ref{sec:hybrid_technique}'de belirtilen şekilde oluşturulmuştur. İlk5 listesi oluşturulurken 5. ürün önerisi olarak İlk5 listesinde daha başarılı olan İTF sistemi kullanılmıştır. Melez sistem başarımları diğer tekil sistemlerin başarımlarına göre önemli bir artış göstermiştir.

\section{Vargılar}
\label{sec:vargilar}

Öneri sistemleri arasında  birbirinden farklı iki önemli yöntem olan İTF ve İF yöntemleri turizm sektöründe otel rezervasyon verisi kullanılarak uygulanmıştır. Yöntemler, performansların karşılaştırılabileceği farklı senaryolar ile test edilmiş ve elde edilen gözlemler raporlanmıştır. 

Sıralı liste birleşimi sayesinde oluşturulan melez öneri sistemi sayesinde öneri performansı tekil sistemlere göre önemli derecede yükseltilmiştir. Müşterilere sunulmak üzere makul uzunlukta olan İlk10 öneri listeleri ele alındığında ve eğitim-test kümesi için tüm verilerin kullanıldığı Senaryo 2'de duyarlılık \%21.46'dır. 

İlerleyen dönemlerde öneri sisteminin devreye alınması ve müşterilere önerilerin bildirilmesi veya gösterilmesi planlanmaktadır. Bu sayede öneri sisteminin daha gerçekçi duyarlılık değerleri alınmış olacaktır. Sunulan öneriler müşteri ilgisini çekerek listeden bir ürün ile etkileşimin gerçekleşme ihtimalinin de artması beklenmektedir.

\section{B{\footnotesize İ}lg{\footnotesize İ}lend{\footnotesize İ}rme}
\textit{Bu çalışma, Etstur tarafından desteklenmiştir.}

\bibliographystyle{IEEEtran}
\bibliography{recommendation}

\end{document}